\documentclass[letterpaper, 10 pt, conference]{ieeeconf}  % Comment this line out if you need a4paper

\IEEEoverridecommandlockouts                              % This command is only needed if 
\usepackage{graphics} % for pdf, bitmapped graphics files
\usepackage{epsfig} % for postscript graphics files
\usepackage{amsmath} % assumes amsmath package installed
\usepackage{amssymb}  % assumes amsmath package installed

\usepackage{amsthm}
\theoremstyle{definition} % Sets the style to 'definition' (bold title, Roman body)
\newtheorem{definition}{Definition}[section] % Defines the environment and how it's numbered
\usepackage{bm}
\usepackage[ruled,lined,linesnumbered]{algorithm2e}

\usepackage{censor}
\usepackage{booktabs}

\usepackage{hyperref}

\newcommand{\etal}{et al.}

\title{\LARGE \bf
Foresight Residual RL for Long-Horizon Robot Manipulation with Vision-Language-Action Models}

\author{Yuhan Liu, Xinyu Zhang, Litao Liu, Abdeslam Boularias
 \thanks{The authors are with the Department of Computer Science, Rutgers
University. This work is supported by NSF award 2132972.}
\thanks{Project website: \href{https://jaysparrow.github.io/foresight-residual-rl}{jaysparrow.github.io/foresight-residual-rl}.}
}
\renewcommand{\eqref}[1]{Eq.~\ref{#1}}
\begin{document}
\setlength{\textfloatsep}{4pt}% Remove \textfloatsep

\maketitle
\thispagestyle{empty}
\pagestyle{empty}

%%%%%%%%%%%%%%%%%%%%%%%%%%%%%%%%%%%%%%%%%%%%%%%%%%%%%%%%%%%%%%%%%%%%%%%%%%%%%%%%
\begin{abstract}
Vision-Language-Action (VLA) policies offer strong general-purpose manipulation priors, but often fail on tight-tolerance, contact-rich assembly due to long-horizon credit assignment and \emph{subtask coupling}: a state that is geometrically successful for the current skill can be brittle for downstream skills. We show this failure mode in residual reinforcement learning (RL) over a frozen VLA base policy: constant sparse success rewards improve each subtask in isolation yet yield little or no gain when skills are chained, because terminal state \emph{quality} is uncontrolled.
We propose \textbf{Foresight Residual RL}, which optimizes handoff quality by augmenting each subtask's sparse success reward with an offline-estimated \emph{foresight value}---the probability of future subtask success conditioned on the  terminal state of the current subtask. Concretely, we (i) train a visual foresight predictor from images of terminal states of the base policy, labeled using downstream rollout statistics, and (ii) train residual policies via backward foresight induction, using the predictor output as a reward multiplier.
On a three-phase wrench-based nut-tightening assembly task in Isaac Gym (grasp, move-insert, rotate), our method achieves \textbf{85.6\%} full-task success, outperforming standard subtask residual RL (54.5\%) and VLA baselines, while leaving per-subtask success unchanged. These results highlight that improving long-horizon performance requires shaping \emph{which} successful states are produced at each sub-task, not only whether success occurs.
\end{abstract}

%%%%%%%%%%%%%%%%%%%%%%%%%%%%%%%%%%%%%%%%%%%%%%%%%%%%%%%%%%%%%%%%%%%%%%%%%%%%%%%%
% =============================================================================
% Introduction
% =============================================================================
\vspace{-0.3cm}
\section{Introduction}
\label{sec:intro}

Many manipulation tasks of practical interest, including industrial assembly, tool use, and surgical manipulation, require executing a sequence of contact-rich skills over long horizons. The standard practice is to decompose such tasks into sub-tasks, train a policy for each sub-task, and chain them together at execution time. This strategy relaxes the challenging long-horizon learning problem to manageable horizons, enables skill reuse with pre-trained policies such as Vision-Language-Action (VLA) 
models~\cite{black2024pi0, kim2024openvla, octo2024}, and allows independent improvement for each sub-task.

However, many long-horizon manipulation problems can not be decomposed into independent sub-tasks. In  assembly operations, the robot must \emph{actively maintain} the success state of a previous sub-task while executing the next. Consider tightening a nut with a wrench: after grasping the wrench and inserting it onto the nut, the robot must keep the wrench inserted during rotation, a condition that is not passively preserved by gravity or geometry but must be actively sustained against contact forces. This \emph{causal coupling} between sub-tasks distinguishes our setting from existing long-horizon manipulation benchmarks, where subtask outcomes are automatically stable once achieved, such as placing an object on a shelf, or opening a drawer~\cite{robocasa2024, heo2023furniturebench}.

The coupling between sub-tasks leads to a consequential failure mode. When each sub-task policy is trained independently with a constant reward upon success, all successful terminal states for the current sub-task are treated as equally valuable for the final task success. In practice, some states, despite being successful for the current sub-task, are difficult to manipulate further and can easily lead to future failures. 
For example, an infirm grasp, despite satisfying the geometric success criterion, leaves the wrench in a fragile configuration that fails during insertion; an insertion that is kinematically valid but shallow will not survive rotation forces. The quality of a sub-task's terminal state, as measured by the probability of future success, varies substantially across the space of geometrically successful completions, yet this variation is neglected in the standard practice.

\begin{figure}[tp]
    \centering
    \includegraphics[width=1\linewidth]{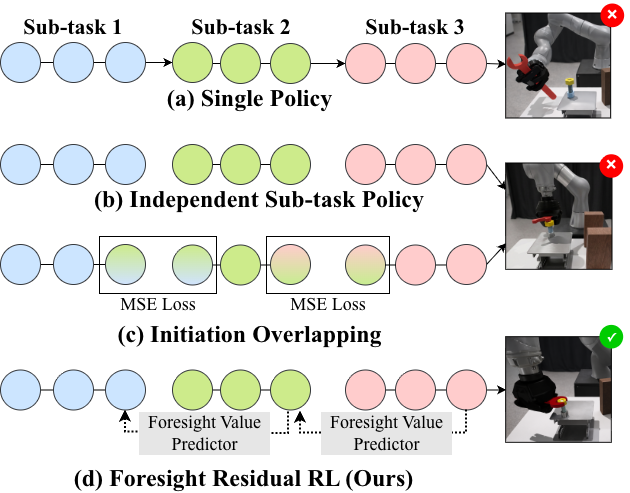}
    \vspace{-0.6cm}
    \caption{\textbf{Existing Works versus Our Method.} 
    Single Policy learns the entire long-horizon task directly. Independent Sub-task  Policy does not consider the dependencies between subtasks. Initiation Overlapping aligns adjacent skills by distribution matching. 
    Our Foresight Residual RL provides a principled coupling between global goals and local updates through foresight value predictor.
    }
    \label{fig:method-compare}
\end{figure}

Prior works on sequential skill composition can be broadly categorized by how it handles this dependency between sub-tasks, as shown in Fig.~\ref{fig:method-compare}. The first category trains a single policy to solve multiple sub-tasks as a single task~\cite{mees2022calvin}. However, these methods can only solve simple manipulation tasks such as pushing buttons or pulling a drawer.  
The second category trains one policy for each sub-task in isolation with task-specific rewards, ignoring inter-task dependencies. This includes standard VLA fine-tuning~\cite{black2024pi0, kim2024openvla}, residual reinforcement learning on individual subtasks~\cite{ankile2024resip, silver2018residual, johannink2019residual}, and imitation-based skill libraries~\cite{mandlekar2023mimicgen, dalal2024planseqlearn}. These methods can produce strong individual skills but do not optimize for the coupling between sub-tasks.
The third category addresses the sub-task coupling issue through distribution matching: regularizing the terminal state distribution of one skill to overlap with the initial states of the next. This category is often referred to as \textit{initiation overlapping}. Adversarial skill chaining~\cite{lee2021tstar}, dual regularization~\cite{chen2024scar}, transition feasibility functions~\cite{chen2023sequential}, and learned initiation sets~\cite{bagaria2023initiation, bagaria2021composable} all fall in this category. Despite their impressive results, these methods only optimize a \emph{proxy}, i.e., distributional overlap, rather than the final task success. A terminal state may lie within the next skill's initial set yet still yield low performance due to the bias of policies in future sub-tasks.

Our key observation is that the \emph{foresight value function}, defined as the expected future return conditioned on the previous sub-task’s terminal state, is exactly the information discarded by constant-reward training and only loosely captured by distribution matching. Notably, the foresight value function has a well-known theoretical counterpart called the completion function in the MAXQ framework~\cite{dietterich2000maxq} and reward-respecting sub-tasks~\cite{sutton2023reward}.
The MAXQ framework~\cite{dietterich2000maxq} learns completion functions that capture future value and reward machines~\cite{icarte2022reward} expose cross-phase reward structure. 
These approaches are theoretically sound but difficult to be deployed in practical applications because they require all sub-task policies to be trained simultaneously, which is not suitable for long-range credit assignment and has poor sample complexity.
Moreover, none of these existing methods have been utilized for modern image-based VLA frameworks. We propose to estimate the foresight value offline and use it to directly augment the sub-task reward during residual RL training on a VLA base policy.

Specifically, our method proceeds in two stages at each sub-task boundary. First, we train a \emph{visual foresight predictor} that estimates the next sub-task's success probability from the current sub-task's terminal observation (image). The predictor is trained offline: we collect terminal states from the base VLA policy, evaluate each by rolling out the downstream policy and distill the resulting success statistics into a probabilistic predictor via binomial maximum likelihood. Second, we train residual policies via \emph{backward foresight induction}: starting from the last phase and working backward, each subtask's residual policy is trained with the foresight predictor's output as a reward multiplier on the sparse local reward signal, steering the policy toward terminal states that benefit the future sub-tasks' success.

We evaluate the proposed method on a highly challenging wrench-screw assembly task in an Isaac Gym simulation, comprising three causally-coupled phases: grasp, move-insert, and rotate. Our results show that our foresight residual RL does not significantly improve individual subtask success rates, the residual policies achieve geometric success at comparable rates with or without foresight, but it significantly improves full-task success when the subtasks are chained. This outcome directly validates the method's purpose: it optimizes \emph{which} successful states are produced, not \emph{whether} subtask success is achieved, closing the gap between isolated subtask performance and composed task performance.

Our contributions are:
\begin{enumerate}

\item We propose Foresight Residual Reinforcement Learning, a novel framework that addresses the causal dependencies between sub-tasks in long-horizon sequential manipulation. By optimizing a sub-task’s terminal state using its foresight value, defined as the expected future return, our approach theoretically aligns local skill execution with global task success.

\item We introduce a practical and scalable implementation of this framework for Vision-Language-Action (VLA) models. We develop  a stage-wise training paradigm comprising offline visual foresight learning and backward foresight induction, enabling efficient end-to-end residual RL over the $\pi_0$ model.

\item We evaluate our method on a challenging, contact-rich wrench-screw assembly task. Our  results demonstrate that foresight residual RL significantly improves the success rate against standard residual RL (+31.1\%), effectively addressing the sub-task coupling problem. 
\end{enumerate}

\begin{figure*}[tp]
    \centering
    \includegraphics[width=1\linewidth]{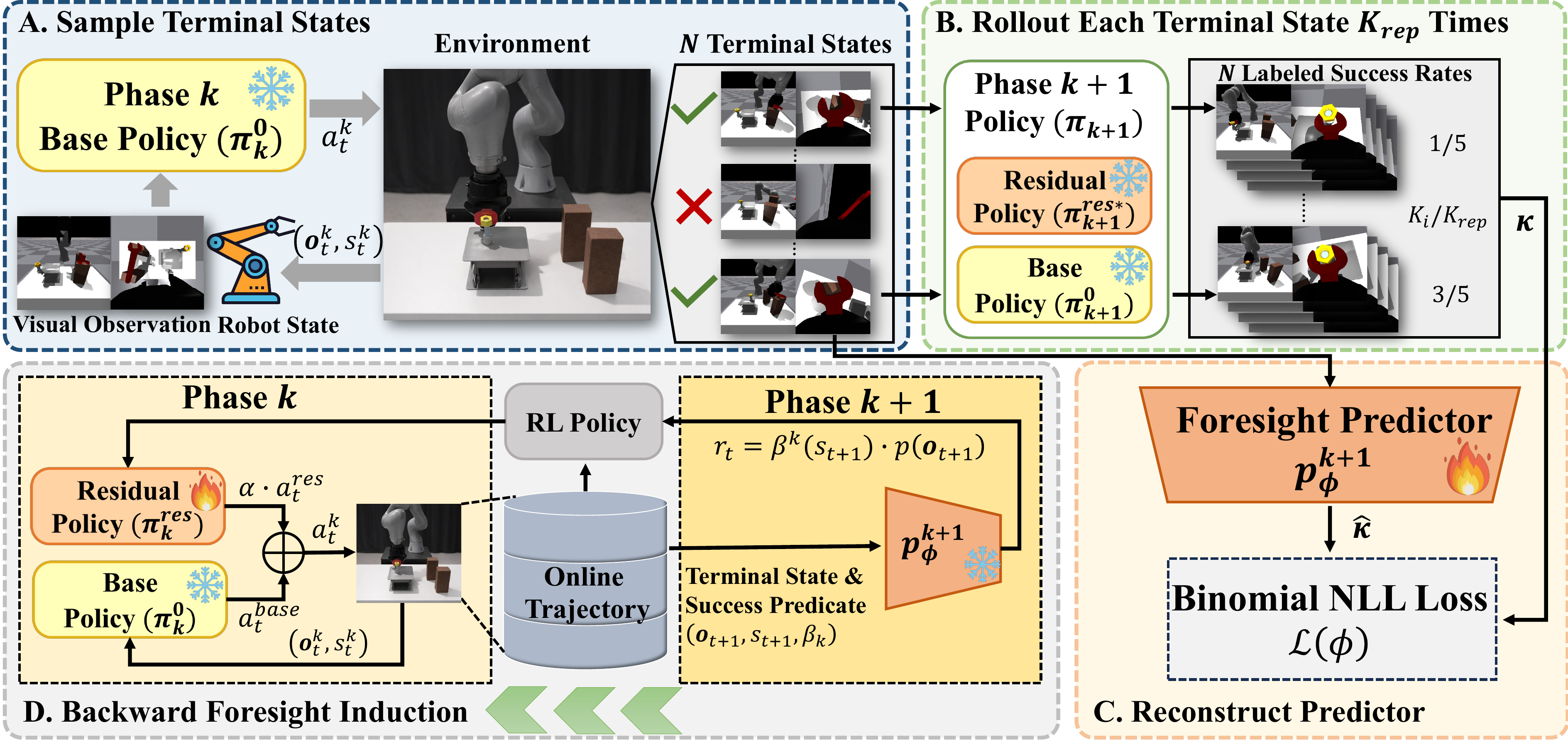}
    \vspace{-0.6cm}
    \caption{\textbf{Overview of Foresight Residual RL.}
    \textbf{(A)} Sample $N$ terminal states at sub-task $k$ by rolling out the frozen base policy.
    \textbf{(B)} For each terminal state, perform $K_{\mathrm{rep}}$ rollouts of sub-task $k{+}1$ to estimate empirical success rates.
    \textbf{(C)} Train a foresight predictor $p_{\phi}^{k+1}$ with a binomial negative
log-likelihood objective to map terminal states of sub-task $k$ to success probabilities of sub-task $k+1$ (Alg.~\ref{alg:predictor}).
    \textbf{(D)} Apply backward foresight induction (Alg.~\ref{alg:pipeline}) to convert predicted foresight values into shaped rewards, and train the residual policy of sub-task $k$  to align it with global task success.}
    \label{fig:pipeline}
    \vspace{-0.5cm}
\end{figure*}

\section{Related Work}
\label{sec:related}

\textbf{Residual RL for manipulation.}
Residual policy learning~\cite{silver2018residual, johannink2019residual} trains a corrective policy on top of a fixed base controller, preserving the base policy's capabilities while improving task-specific performance via RL.
Recent work extends this paradigm to vision-based imitation learning policies: ResiP~\cite{ankile2024resip} trains per-time-step residual corrections on frozen diffusion policies for precision assembly; ResFiT~\cite{ankile2025resfit} achieves sample-efficient residual fine-tuning via off-policy RL; and PLD~\cite{xiao2025pld} applies residual RL to VLA models including $\pi_0$~\cite{black2024pi0} with a probe-learn-distill cycle.
These methods train residual policies for \emph{individual} tasks with standard reward functions. Our work inherits the residual RL formulation but introduces foresight-corrected rewards that account for downstream phase success, a cross-phase optimization objective that none of these methods consider.

\textbf{Skill sequencing and transition optimization.}
Composing independently trained skills into long-horizon behaviors is a longstanding challenge~\cite{konidaris2009skill}.
A central difficulty is that the terminal state distribution of one skill may not align with the effective initiation set of the next.
T-STAR~\cite{lee2021tstar} addresses this via adversarial regularization of terminal states; SCaR~\cite{chen2024scar} adds dual regularization across skill boundaries; Sequential Dexterity~\cite{chen2023sequential} learns transition feasibility functions for bi-directional optimization. Bagaria~\etal~\cite{bagaria2023initiation, bagaria2021composable} learn robust initiation set classifiers.
Lee~\etal~\cite{lee2019transition} take a different approach, inserting dedicated transition policies between skills, while Meehan\etal~\cite{meehan2024composing} adapt option policies to bridge distribution gaps.
All of these methods optimize a \emph{distributional proxy}: they seek overlap between one skill's output and the next skill's input, without directly measuring whether that overlap leads to downstream task success. Our foresight predictor replaces this proxy with the next phase's success probability, and uses it as a reward signal rather than a distributional constraint.

\textbf{Value decomposition in hierarchical RL.}
The theoretical foundations for cross-phase value propagation are well established.
MAXQ~\cite{dietterich2000maxq} decomposes hierarchical value functions via \emph{completion functions} that capture expected return after subtask termination.
The options framework~\cite{sutton1999options} propagates value across option boundaries through $V_\Omega(s')$ at termination states, and option-critic~\cite{bacon2017option} derives end-to-end gradients through these boundaries.
Sutton~\etal~\cite{sutton2023reward} formalize \emph{reward-respecting subtasks} whose cumulants incorporate the task reward plus a terminal bonus, ensuring subtask optimization accounts for the parent task's objective.
Our foresight value function is a direct instantiation of the MAXQ completion function restricted to a linear task graph. The key difference is operational: rather than learning completion functions online via temporal-difference updates within a joint hierarchy, which introduces non-stationarity and precludes the use of frozen pre-trained base policies, we estimate the foresight value \emph{offline} via Monte Carlo rollouts of fixed downstream policies and distill it into an image-based predictor. This decouples foresight estimation from sub-task policy training, enabling a single-pass backward induction procedure compatible with residual RL on VLA base policies.

\section{Problem Formulation}
\label{sec:problem}

\subsection{Sequential Semi-MDP}

We consider a long-horizon manipulation task defined as an MDP $\mathcal{M} = (\mathcal{S},  \mathcal{A}, T, R, \gamma)$ that admits a decomposition into $K$ sequential \emph{phases} $\{1, \ldots, K\}$. Each phase $k$ represents a sub-task and has a termination predicate $\beta_k : \mathcal{S} \to \{0, 1\}$ indicating sub-task success. Sub-tasks are performed in order: phase $k+1$ begins from the terminal state of phase~$k$. We assume the success predicates $\{\beta_k\}_{k=1}^K$ are available to all phases at any time.

This structure forms a sequential semi-MDP where phases serve as temporally extended actions.
Unlike typical option frameworks, our phases are \emph{causally coupled}: phase $k+1$ must actively maintain the success conditions established by phase $k$ while pursuing its own objective (e.g., keeping the wrench inserted while turning the nut).

Our goal is to learn a policy $\bm{\pi}$ that maximizes the expected return of the \emph{final} phase success reward $\beta_K$:
\begin{equation}
    V(s) = \mathbb{E}_{\bm{\pi}}\left[\sum_t \gamma^t\beta_K(s_{t+1})\Big|s_0=s\right],
\end{equation}
where $\beta_K$ is the single source of the reward signals. $\beta_K$ indicates whether the final sub-task has succeeded or not. Note that this can happen before the end of the episode, and last for a number of time-steps, hence the use of the discounted sum.

For a long-horizon manipulation task with a sparse reward, end-to-end RL struggles to learn an effective policy. Therefore, a common practice is to decompose the whole task into phases and learn an individual policy $\pi_k$ for each phase $k$. The final policy is the composition of these policies into a sequence $\bm{\pi} = (\pi_1, \ldots, \pi_K)$.

\subsection{Task Decomposition}
\label{sec:standard_decomposition}

The standard approach in the literature is to train a policy $\pi_k$  for each phase $k$ independently using a \emph{constant} terminal reward $r(s_t, a_t)=\beta_k(s_{t+1})$, which is positive upon subtask completion. The objective is then to maximize the local value function $V_k$ for each phase $k$:
\begin{equation}
    V_k(s) = \mathbb{E}_{\bm{{\pi_k}}}\left[\sum_{t=0}^{\bm{T_k}} \gamma^t\beta_{\bm{k}}(s_{t+1})\Big|s_0^{(k)}=s\right],
    \label{eq:standard_value}
\end{equation}
where $T_k$ is the max time-steps reserved for phase $k$ and $s_0^{(k)}$ is the initial state of phase $k$.
Typically, success reward $\beta_k$ does not take into account whether the next sub-task $k+1$ will be successful or not. It only considers the success of sub-task $k$. 
However,  in long-horizon robotic manipulation tasks, success in latter phases often depends on the policy's performance in earlier phases. 
Therefore, learning policies by maximizing $V_k(s)$ for each phase $k$ independently, using only sub-task success predicates $\beta_k$ as reward, leads to sub-optimal performance.

\section{Foresight Residual RL}
\label{sec:method}

\subsection{Foresight Value Function}
\label{sec:foresight_value}

\begin{definition}[Foresight value function]
\label{def:foresight}
For phase $k < K$ with fixed future policies $\pi_{k+1:K}$, the foresight value function is the expected discounted final-phase success starting from state $s$ at the beginning of phase $k+1$:
\begin{equation}
    V_{k+1:K}(s) \;=\; \mathbb{E}_{\pi_{k+1:K}}\!\left[\sum_{t=0}^{T_{k+1:K}} \gamma^{t}\,\beta_K(s_{t+1}) \;\Big|\; s_0^{(k+1)} = s \right],
    \label{eq:foresight_corrected}
\end{equation}
where $T_{k+1:K} = \sum_{j=k+1}^{K} T_j$, the sum of each remaining sub-task's horizon $T_j$.
\end{definition}

$V_{k+1:K}(s)$ captures the future consequence of terminating phase $k$ in state $s$. In doing so, $V_{k+1:K}(s)$ captures the inter-dependent nature of the sub-tasks and the effect that a terminal state of a sub-task can have on subsequent sub-tasks. 
This formulation stems from our observation that in contact-rich assembly, where small pose variations at handoff produce large downstream success differences, any improvement in full-task performance must come from producing terminal states with higher expected foresight value.

\subsection{Local Foresight Value Function}

We multiply the terminal reward $\beta_k$ of the local value function in \eqref{eq:standard_value} with the foresight value function $V_{k+1:K}$, this results in the local foresight value function in \eqref{eq:foresight_objective}, which is then the new learning objective for phase $k$.
\begin{equation}
    V_k^{\circ}(s) = \mathbb{E}_{\pi_k}\!\left[\sum_{t=0}^{T_k} \gamma^t\, \beta_k(s_{t+1}) V_{k+1:K}(s_{t+1})\big|s_0^{(k)}=s\right].
    \label{eq:foresight_objective}
\end{equation}

Compared to the local value function in  \eqref{eq:standard_value}, the only difference is the product of the reward of phase-$k$ success and the state-dependent $V_{k+1:K}(s_{t+1})$ at phase termination. This correction is \emph{local}: it modifies only the terminal reward and preserves the optimization process within each phase.

\subsection{Backward Induction over Phases}

\begin{algorithm}[t]
\caption{Backward Foresight-driven Learning}
\label{alg:pipeline}
\KwIn{Base policies $\{\pi_k^0\}_{k=1}^K$, residual scale $\alpha$}
\KwOut{Trained residual policies $\{\pi_k^{\text{res}}\}_{k=1}^K$}
\For{$k = K, K{-}1, \ldots, 1$}{
    \eIf{$k = K$}{
        Train $\pi_K^{\text{res}}$ via PPO with reward $r_t = \beta_K(s_{t+1})$ and actions $\mathbf{a}_t \sim (\pi_k^0(\mathbf{o}_t) + \alpha \cdot \pi_k^{\text{res}}(\mathbf{o}_t))$\;
    }{
        $\pi_{k+1}(\bm{o})\gets\pi_{k+1}^0(\bm{o})+\alpha\cdot \pi_{k+1}^{\text{res}}(\bm{o}), \forall \bm{o}$\;
        \tcp{$\bm{o}$ is an image observation}
        $p_\phi \gets \textsc{ForesightPredictor}(\pi_k^0,\; \pi_{k+1})$\;
        Train $\pi_k^{\text{res}}$ via PPO with reward $r_t =  \beta_k(s_{t+1}) \cdot p_\phi(\mathbf{o}_{t+1})$
        and actions $\mathbf{a}_t \sim (\pi_k^0(\mathbf{o}_t) + \alpha \cdot \pi_k^{\text{res}}(\mathbf{o}_t))$\;
        
    }
}
\end{algorithm}

\begin{algorithm}[t]
\caption{\textsc{ForesightPredictor}}
\label{alg:predictor}
\KwIn{Base policy $\pi_k^0$, downstream policy $\pi_{k+1}$, number of episodes $N$ and rollouts $K_{\text{rep}}$}
\KwOut{Foresight predictor $p_\phi$}
\tcp{Stage 1: Collect terminal states from base policy}
$\mathcal{D} \gets \emptyset$\;
\For{$i = 1, \ldots, N$}{
    Roll out $\pi_k^0$ until $\beta_k(s) = 1$ or horizon is reached\;
    \If{$\beta_k(s) = 1$}{
        Record terminal observation $\mathbf{o}_i$ and state $s_i$\;
    }
}
\tcp{Stage 2: Label terminal states with future success}
\For{each collected $(s_i, \mathbf{o}_i)$}{
    \For{$j = 1, \ldots, K_{\emph{rep}}$}{
        Roll out $\pi_{k+1}$ from $s_i$; record $z_j = \beta_{k+1}(s_{\text{final}})$ where $s_{\text{final}}$ is the final state in the rolled out trajectory\;
    }
    $\kappa_i \gets \sum_{j=1}^{K_{\text{rep}}} z_j$\;
    $\mathcal{D} \gets \mathcal{D} \cup \{(\mathbf{o}_i,\; \frac{\kappa_i}{K_{\text{rep}}})\}$\;
}
\tcp{Stage 3: Train the predictor}
Optimize $p_\phi$ on $\mathcal{D}$ with binomial negative log-likelihood (Eq.~\ref{eq:binomial_loss})\;
\Return{$p_\phi$}
\end{algorithm}

Since foresight value function $V_{k+1:K}$ depends on future phases, we propose to train sub-task policies in their reverse order (Alg.~\ref{alg:pipeline}). For $k = K, K{-}1, \ldots, 1$: if $k = K$, the foresight value is set to $V_{K+1:K}\triangleq1$ and we train $\pi_K$ with the standard task reward (no foresight needed for the last phase); if $k < K$, 
% fix all future policies $\pi_{k+1:K}$, 
we estimate $V_{k+1:K}$ by evaluating $\pi_{k+1:K}$ from phase-$k$ terminal states (Sec. \ref{sec:foresight_estimation}), and train $\pi_k$ with the local foresight objective~\eqref{eq:foresight_objective}.

The backward ordering is essential because training sub-task $k$ requires rolling out the
downstream policies $\pi_{k+1:K}$ to estimate $V_{k+1:K}$, so these policies must
already be trained and fixed. Because the downstream policies do not change during
sub-task $k$ training, the foresight estimates are stationary, unlike in joint
hierarchical optimization methods where all levels update simultaneously.

\section{Learning Foresight Value Functions}
\label{sec:practical}

\subsection{Residual Policy for Terminal State Correction}

Let $\pi_k^0$ be the base VLA policy for sub-task $k$. The residual policy $\pi_k^{\text{res}}$ outputs small corrections:
\begin{equation}
    \mathbf{a}_t = \pi_k(\mathbf{o}_t) \triangleq \pi_k^0(\mathbf{o}_t) + \alpha \cdot \pi_k^{\text{res}}(\mathbf{o}_t), \qquad \alpha \ll 1,
    \label{eq:residual_action}
\end{equation}
where $\mathbf{o}_t$ is the observations at step $t$ and $\alpha$ is a small weight of the residual policy in the final action.

Using the learned foresight values to train the residual policy, the residual policy shifts the base policy's terminal states at phase $k$ toward regions of higher foresight value. 
\vspace{-2mm}
\subsection{Foresight Value Estimation}
\label{sec:foresight_estimation}

Computing $V_{k+1:K}(s)$ requires evaluating the downstream return from state $s$. 
A direct approach is to estimate $V_{k+1:K}$ on-the-fly during RL training of phase $k$: at each encountered terminal state of sub-task $k$, spawn rollouts of $\pi_{k+1:K}$ from that state, collect the Monte Carlo return, and use it as the terminal reward. This produces unbiased estimates of the foresight value $V_{k+1:K}(s)$ but has two significant drawbacks.
First, the downstream return has high variance due to the stochasticity of both $\pi_{k+1:K}$ and the environment dynamics, requiring many rollouts per terminal state to obtain reliable estimates. This variance propagates directly into the RL training signal, destabilizing policy gradient updates.
Second, each terminal state evaluation requires executing full downstream episodes in simulation, incurring substantial computational overhead that scales with the number of remaining sub-tasks and their horizons. In our setting of the wrench-screw task, evaluating a single move-insert terminal state requires rolling out up to 200 steps of the rotate policy, multiplied by the required repetitions.\\
\indent A further alternative is to reuse the downstream policy's own value critic as the foresight estimate. However, that critic is calibrated on the downstream policy's initial-state distribution, whereas the reward must score the upstream policy's terminal handoff states. Our predictor is instead fit directly on these terminal states with Monte-Carlo success labels, yielding calibrated estimates exactly where they are applied.\\
\indent\textbf{Our approach: Offline value estimation with a learned predictor.}
Our instantiation consists of a one-step foresight approximation and an offline estimation procedure.\\
\indent\emph{1) One-step foresight approximation.} We approximate the full downstream value by the immediate next sub-task's success probability:
\begin{equation}
\begin{aligned}
        V_{k+1}(s) &= \Pr\!\big(\text{sub-task } k{+}1 \text{ succeeds} \mid s_0^{(k+1)} = s,\; \pi_{k+1}\big)\\ &\approx V_{k+1:K}(s).
\end{aligned}
    \label{eq:one_step}
\end{equation}
This is exact when: (a) $k = K{-}1$ (the next phase is the last one), or (b) the success rate of sub-tasks $k{+}2, \ldots, K$ is constant conditioned on sub-task $k{+}1$ succeeding. When deeper foresight is needed, this approximation can be extended by chaining rollouts through all remaining phases.

\emph{2) Offline data collection and predictor training.}
Following Alg.~2, we roll out the base policy $\pi_k^0$ (without residual)
to collect $N$ successful terminal states, execute $K_{\text{rep}}$
independent rollouts of $\pi_{k+1}$ from each, and record the success
count $\kappa_i$ to form the label $y_i = \kappa_i / K_{\text{rep}}$ for
each terminal observation $o_i$.

Using the base policy's distribution rather than the augmented policy's (base+residual) is deliberate: the predictor needs to be accurate on the states the residual will visit \emph{before} training shifts them toward better configurations.

Compared to online Monte Carlo, this approach amortizes the cost of downstream evaluation into a one-time offline data collection step. The learned predictor then provides instant, deterministic foresight estimates at RL training time, eliminating both the per-step simulation overhead and the variance of sampled returns.

\subsection{Visual Foresight Predictor}

We distill the Monte Carlo estimates into a neural predictor $p_\phi : \mathcal{O} \to [0, 1]$, defined as
\begin{equation}
    p_\phi(\mathbf{o}) \approx V_{k+1}(s),
\end{equation}
where $\mathbf{o}$ is the observation (image) at the terminal state.

\textbf{Training objective.} Since each label $y_i$ derives from $\kappa_i$ successes in $K_{\text{rep}}$ trials, we train with the binomial negative log-likelihood:
\begin{equation}
    \mathcal{L}(\phi) = -\sum_{i=1}^{N}\Big[\kappa_i \log p_\phi(\mathbf{o}_i) + (K_{\text{rep}} - \kappa_i)\log\!\big(1 - p_\phi(\mathbf{o}_i)\big)\Big].
    \label{eq:binomial_loss}
\end{equation}
This is the maximum likelihood estimator for the Bernoulli parameter of a binomial observation model. It respects the count structure of the labels rather than treating $y_i$ as a regression target, yielding calibrated probabilities when the model class is sufficiently expressive.

\textbf{Distribution shift under policy improvement.}
As the residual policy improves, it visits terminal states not seen during the  training of the predictor. The small residual scale $\alpha$ limits this shift in practice, keeping the augmented policy's terminal distribution close to the base policy's and within the interpolation range of the predictor.

\subsection{Foresight-Modulated Reward}

Since the only reward signal comes from the success predicates, the training reward for phase $k$ at step $t$ is:
\begin{equation}
    r_t = \beta_k(s_{t+1}) \cdot p_\phi(\mathbf{o}_{t+1}),
    \label{eq:foresight_reward}
\end{equation}
The predictor $p_\phi$ serves as a plug-in estimator for $V_{k+1:K}$. This is the \emph{only} reward the agent receives, there is no additional shaping or internal reward.
\section{Experiments}
\label{sec:experiments}

\subsection{Task Setup}

We evaluate our method on a wrench-screw assembly task implemented in NVIDIA Isaac Gym~\cite{makoviychuk2021isaacgym} with GPU-accelerated PhysX contact simulation. A 7-DoF Kuka IIWA arm with an 11-DoF Robotiq 3-finger gripper must grasp a wrench from a holder, transport and insert it onto a hex nut, and rotate the nut counterclockwise by at least $60^\circ$ while maintaining insertion. The task decomposes into three phases: \textsc{Grasp} (160 steps), \textsc{Move-Insert} (200 steps), and \textsc{Rotate} (200 steps), for a maximum episode length of 560 steps at 20\,Hz control frequency.

The learned policies observe two RGB images (third-person and wrist camera, $256 \times 256$) and 8D proprioception (end-effector pose + binary gripper state). No force feedback or ground-truth object poses are available to policies. Actions are 7D operational-space-control deltas (3D position, 3D rotation in axis-angle, 1D gripper command).

At each episode reset, the initial poses of the wrench holder, nut, and bolt are randomized. We use 5 discrete wrench-nut size variants with less than 2\,mm clearance between the wrench head and nut, requiring sub-millimeter precision for insertion. All randomizations are identical across methods.

\subsection{Training Pipeline}
\label{sec:training}
\textbf{Expert demonstrations.}
We collect 2{,}560 demonstrations per sub-task with a state-machine expert that uses privileged information unavailable to the learned policies, consisting of only ground-truth object poses, proprioception, and known asset geometry. For insertion, the expert is a pre-trained state-based RL policy observing noisy nut-frame poses \cite{liu2025failure}. The demonstrations are homogeneous: the expert follows a fixed strategy per sub-task and does not exhibit recovery behaviors, producing a narrow terminal state distribution.

\textbf{Base VLA policy.}
We fine-tune the generalist $\pi_0$ policy~\cite{black2024pi0} separately on each sub-task's demonstration data, yielding three per-sub-task base policies. Each policy uses action chunking with horizon $H = 20$, re-planning only when the action buffer is exhausted or a phase transition occurs.

\textbf{Residual RL policy.}
Whereas the base policy conditions on both camera views and proprioception, the residual policy inputs only the wrist
view: a frozen DINOv2-S/14 backbone extracts $16\times16$ patch
features from the wrist camera image, followed by a CNN ($384\to12$ channels), a 2-layer LSTM (256 hidden units), and an MLP ($512\to256\to128$) with separate actor and critic heads. The actor output layer is zero-initialized so that the residual produces near-zero corrections at the start of training, preserving the base policy's behavior. Residual scale is $\alpha = 0.1$.

We train with PPO~\cite{schulman2017ppo} via rl\_games~\cite{rl-games2021} across 128 parallel environments. Training proceeds in two phases: a 10-iteration critic-only warmup (actor frozen, executing only the base policy) to adapt the value function to the base policy's performance distribution, followed by 300 iterations of joint actor-critic training. Phase transitions during training use flag-based (reactive) detection: \textsc{Grasp}$\to$\textsc{Move-Insert} triggers when the wrench-grasped predicate activates; \textsc{Move-Insert}$\to$\textsc{Rotate} triggers when wrench-inserted activates.

\subsection{Foresight Predictor}

Following Algorithm~\ref{alg:pipeline}, we train one predictor per phase boundary (grasp$\to$insert and insert$\to$rotate). For each boundary, we collect $N = 2{,}560$ successful terminal states from the base VLA policy and evaluate each by executing $K_\text{rep} = 5$ independent rollouts of the downstream policy (base + trained residual), recording per-state success counts $\kappa_i$.

The foresight value predictor uses the DINOv2-S/14 CLS tokens from both camera views, fused via self-attention and a linear layer, and mapped to a scalar output through a sigmoid head. The weights of the DINO backbone are finetuned using LoRA~\cite{hu2021lora} with a rank of 8. We train with binomial NLL (Eq.~\ref{eq:binomial_loss}) using an 80/10/10 train/validation/test split for 20 epochs (less than 5 minutes).

% =============================================================================

\section{Results}
\label{sec:results}

\subsection{Foresight Predictor Quality}

\begin{table}[t]
    \centering
    \caption{Foresight predictor evaluation on held-out test sets.}
    \label{tab:predictor}
    \begin{tabular}{lcc}
        \toprule
        Boundary & Prediction Accuracy & Dataset Success Rate \\
        \midrule
        Grasp $\to$ Insert    & $86.2\%$ & $73.9\%$\\
        Insert $\to$ Rotate   & $93.9\%$ & $90.4\%$ \\
        \bottomrule
    \end{tabular}
\end{table}

Table~\ref{tab:predictor} reports held-out classification accuracy (at threshold $0.5$)
for each boundary predictor, alongside the dataset success rate as a majority-class
baseline. Both predictors exceed their respective baselines, with the larger margin at
the Grasp$\to$Insert boundary ($86.2\%$ vs.\ $73.9\%$) confirming that the predictor
discriminates among grasp terminal states that are kinematically successful but that lead to different outcomes for the subsequent sub-tasks. We note that the predictor is used as a \emph{continuous} reward multiplier (Eq.~\ref{eq:foresight_reward}), not as a binary classifier; thresholded accuracy
is a conservative proxy for its utility.

\subsection{Main Results}
\begin{table*}[t]
    \centering
    \caption{Success rates (\%) on the wrench-screw assembly task. Subtask results are initialized from expert states (oracle). Consecutive and full-task results are end-to-end without privileged initialization. Best results in \textbf{bold}.}
    \vspace{-0.2cm}
    \label{tab:main}
    \begin{tabular}{l ccc cc c}
        \toprule
        & \multicolumn{3}{c}{Subtask (oracle init)} & \multicolumn{2}{c}{Consecutive} & Full Task \\
        \cmidrule(lr){2-4} \cmidrule(lr){5-6} \cmidrule(lr){7-7}
        Method & Grasp & Insert & Rotate & Grasp+Insert & Insert+Rotate & All \\
        \midrule
        $\pi_0$ (end-to-end)
            & ---
            & ---
            & ---
            & ---
            & ---
            & $54.9\pm 5.3\%$ \\
        $\pi_0$ (chained)
            & $87.1\pm 1.4\%$
            & $45.7\pm 6.1\%$
            & $93.4\pm 3.0\%$
            & $37.1\pm 1.5\%$
            & $61.5\pm4.6\%$
            & $41.4\pm 3.3\%$ \\
        $\pi_0$ + Residual
            & $\bm{98.4\pm 0.6\%}$
            & $\bm{92.2\pm 1.1\%}$
            & $99.2\pm 0.6\%$
            & $55.5\pm 3.4\%$
            & $83.4\pm3.2\%$
            & $54.5\pm 3.6\%$ \\
        $\pi_0$ + Residual with Foresight (ours)
            & $95.7\pm 3.3\%$
            & $91.4\pm0.6\%$
            & $\bm{99.8\pm 0.4\%}$
            & $\bm{87.3\pm 1.3\%}$
            & $\bm{91.8\pm 3.3\%}$
            & $\bm{85.6\pm 3.9\%}$ \\
        \bottomrule
    \end{tabular}
    \vspace{-0.5cm}
\end{table*}

We evaluate each method over 512 episodes across 4 random seeds, and report the mean and standard deviation of success rates. We emphasize that the foresight predictor enters only as a training-time
reward and plays no role at evaluation, with evaluation performed on independently randomized episodes. Table~\ref{tab:main} presents results at three granularities: individual subtasks (initialized from oracle states), consecutive
two-phase chains, and the full task. Here, oracle initialization places each sub-task policy at a successful terminal state of the preceding sub-task produced by the state-machine expert (Sec.~\ref{sec:training}). These
single-subtask numbers measure isolated capability.

With oracle initialization, both constant-reward and foresight-modulated residual
policies achieve high subtask success ($>91\%$ on all phases), substantially improving
over chained $\pi_0$ without residual correction. However, constant-reward residual RL
fails to translate these gains to the full task: its $54.5\%$ success rate is not better
than end-to-end $\pi_0$ ($54.9\%$), despite doubling \textsc{Insert} success in isolation
($45.7\% \to 92.2\%$). This confirms that improving subtask success rate without
controlling sub-task terminal distribution  yields no gain under composition.
Foresight prediction recovers this lost value, achieving $85.6\%$ full-task success, over
$30\%$ improvement over the constant-reward variant.

The gains concentrate at the Grasp$\to$Insert boundary, where consecutive success
improves from $55.5\%$ to $87.3\%$ ($+31.8\%$), which is consistent with the predictor's
larger discriminative margin at this transition (Table~\ref{tab:predictor}).
We also observe that naive decomposition without correction is harmful:
chained $\pi_0$ ($41.4\%$) underperforms end-to-end $\pi_0$ ($54.9\%$), as each
uncontrolled handoff introduces compounding state mismatch. Foresight correction
reverses this tradeoff, making the decomposed approach substantially outperform
the monolithic one.

\subsection{Terminal State Quality}

To verify that the foresight correction shifts the terminal state distribution toward higher-quality configurations, we evaluate the foresight predictor's scores on terminal states produced by each method (Table~\ref{tab:terminal}).

\begin{table}[t]
    \centering
    \caption{Mean foresight predictor score on terminal states of the trained policies. Higher scores indicate terminal states more likely to lead to downstream success.}
    \vspace{-0.2cm}
    \label{tab:terminal}
    \begin{tabular}{l cc}
        \toprule
        & \multicolumn{2}{c}{Mean Predictor Score} \\
        \cmidrule(lr){2-3}
        Method & Grasp terminal & Insert terminal \\
        \midrule
        $\pi_0$ (base, no residual) & $0.64$ & $0.97$ \\
        + Residual      & $0.32$ & $0.86$ \\
        + Residual with Foresight (ours)      & $0.73$ & $0.91$ \\
        \bottomrule
    \end{tabular}
    % \vspace{-0.5cm}
\end{table}

Table~\ref{tab:terminal} reveals a counterintuitive finding: constant-reward residual
RL \emph{degrades} mean predictor scores at both boundaries despite improving subtask
success rates (Table~\ref{tab:main}). At the Grasp terminal, the score drops from
$0.64$ to $0.32$---the policy finds grasps that satisfy the success predicate more
reliably but produce configurations that are worse for downstream insertion.
Foresight correction reverses this degradation, achieving the highest Grasp terminal
score ($0.73$) and retaining more of the base policy's Insert terminal quality
($0.91$ vs.\ $0.86$). This directly explains the divergence in
Table~\ref{tab:main}: constant-reward residual RL improves subtasks in isolation
while eroding the compositional value of their terminal states.

\subsection{Qualitative Analysis}

\begin{figure}    
    \centering
    \includegraphics[width=0.75\linewidth]{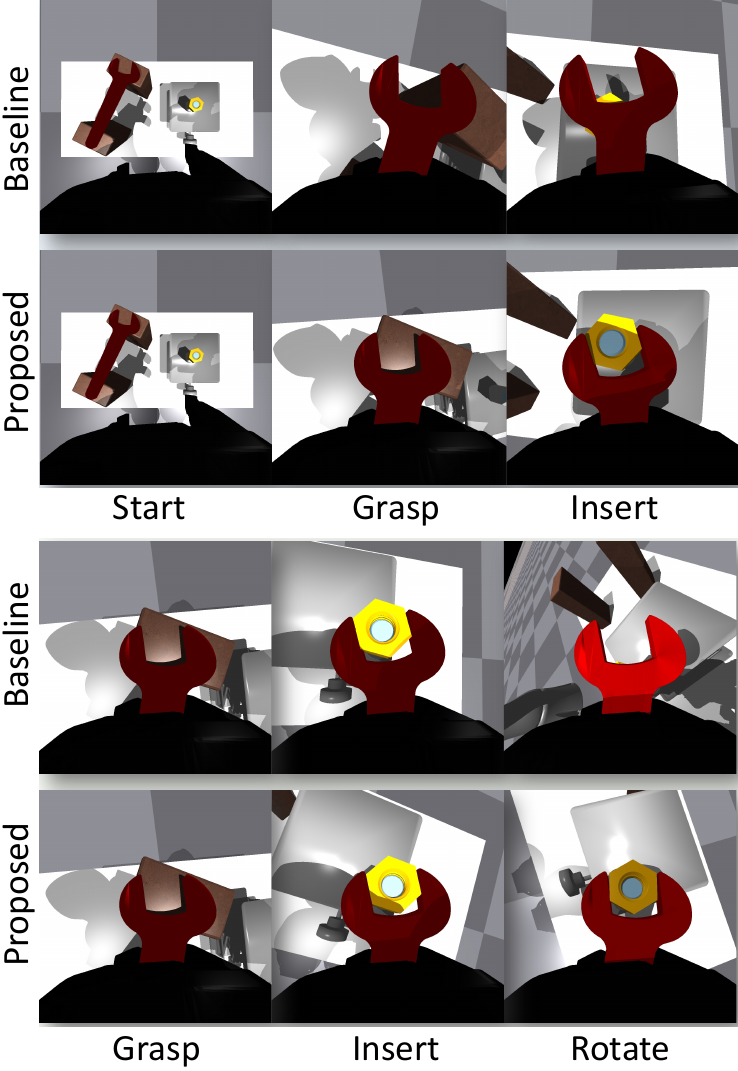}
    \vspace{-0.3cm}
    \caption{Qualitative comparison of terminal states produced by constant-reward
    (baseline) and foresight-modulated (proposed) residual policies. Top: Grasp$\to$Insert
    transition. Bottom: Insert$\to$Rotate transition. Each panel shows, from left to
    right, the shared starting state, the subtask terminal state reached by each method,
    and the outcome of the subsequent phase. Both methods satisfy the subtask success
    predicate (middle), but only foresight-corrected terminal states lead to a downstream
    success.}
    \label{fig:qualitative}
    % \vspace{-0.5cm}
\end{figure}

Fig.~\ref{fig:qualitative} visualizes the mechanism behind the quantitative gap in
Table~\ref{tab:main}. In both panels, the two policies start from identical states
and both achieve the current sub-task. The middle columns show geometrically valid
grasps and insertions respectively, consistent with the comparable sub-task success
rates in Table~\ref{tab:main}. The divergence appears in the immediate next sub-task.

At the Grasp$\to$Insert boundary (top panel), the constant-reward policy grasps the wrench
at an orientation that satisfies the success predicate but provides poor alignment for
transporting it to the nut. When the \textsc{Insert} policy takes over, this misalignment compounds
under contact and the insertion fails (top-right). The foresight residual policy
produces a more centered grasp with better angular alignment, giving the downstream
policy sufficient margin to complete the insertion (bottom-right).

A similar pattern appears at the Insert$\to$Rotate boundary (bottom panel). The
constant-reward policy achieves engagement between the wrench and the nut but at a pose
that lacks stability; when a rotational torque is applied, the wrench
disengages (top-right). The foresight-modulated insertion maintains a tighter
seat, sustaining contact through the rotation phase (bottom-right).

These examples illustrate a general property of causally-coupled sequential
manipulation: the subtask success predicate defines a broad equivalence class
of terminal states, but only a subset composes well with the downstream phase.
Small geometric differences at each handoff, which are invisible to the success
predicate, amplify under the contact dynamics of the subsequent phase.
Foresight correction narrows the terminal distribution toward this composable
subset.

\section{Conclusion}
We introduced \emph{Foresight Residual RL}, a framework for causally-coupled long-horizon manipulation that optimizes \emph{handoff state quality} rather than subtask success alone. Our approach learns an offline visual foresight predictor for each phase boundary and uses it to reweight the subtask success reward for residual policy training over base VLA models. The subtask policies are trained inductively in reverse phase order. On the wrench-screw assembly task, foresight residual RL yields large gains under skill composition: full-task success improves to $85.6\%$, outperforming standard subtask residual RL ($54.5\%$) and the $\pi_0$ baselines. 

\textbf{Limitations and future work.} Our evaluation is conducted entirely
in simulation. We expect the main obstacle to real-world deployment to be
the cost of on-hardware online RL for the residual policy rather than the
contact-dynamics gap, since the base VLA and the offline foresight predictor
transfer through a standard image-based pipeline. We validate on a single wrench-screw task with a known
phase decomposition; additional tasks and decompositions and alternative VLA
backbones are natural next steps.

In conclusion, our results clearly demonstrate that optimizing the terminal state of a subtask, not just \emph{whether} the subtask succeeds, is essential for composing contact-rich manipulation skills.

\bibliographystyle{IEEEtran}
\bibliography{sections/reference} % Entries are in the reference.bib file

\addtolength{\textheight}{-12cm}   % This command serves to balance the column lengths

%%%%%%%%%%%%%%%%%%%%%%%%%%%%%%%%%%%%%%%%%%%%%%%%%%%%%%%%%%%%%%%%%%%%%%%%%%%%%%%%

\end{document}